\documentclass[preprint,12pt]{elsarticle}

\usepackage{amsmath}
\usepackage{booktabs}
\usepackage{url}
\graphicspath{{figures/}{}}

\begin{document}

\begin{frontmatter}

\title{Lesion-centered 3D mapping of colonoscopy procedures: validation of a hierarchical ensemble pipeline on public benchmark videos}

\author[inst1,inst4]{Hyunjun Kim}
\ead{hyunjun1121@kaist.ac.kr}

\author[inst2,inst4]{Hyeonwoo Na}
\ead{real\_na@eis.hokudai.ac.jp}

\author[inst3,inst4]{Jaewoo Lee\corref{cor1}}
\ead{ljw980905@gmail.com}
\cortext[cor1]{Corresponding author.}

\affiliation[inst1]{organization={School of Computing, KAIST}, city={Daejeon}, country={Republic of Korea}}
\affiliation[inst2]{organization={Division of Mechanical and Space Engineering, Faculty of Engineering, Hokkaido University}, city={Sapporo}, country={Japan}}
\affiliation[inst3]{organization={CHA University School of Medicine}, city={Seongnam}, country={Republic of Korea}}
\affiliation[inst4]{organization={Clinical Imaging Research Institute}, city={Seoul}, country={Republic of Korea}}

\begin{abstract}
\textbf{Background and Objective}: Colonoscopy recording practice preserves text reports and still photographs, while the spatial information already present in the recorded video --- where the scope traveled, where a lesion was observed, and whether the same lesion was seen again --- is discarded when the procedure ends. This study determines whether a lesion-centered spatial record can be assembled and validated without full-colon 3D reconstruction. \textbf{Methods}: A four-layer hierarchical pipeline was assembled --- (1) a global topological map, (2) lesion-level spatio-temporal tracks, (3) on-demand local 3D reconstruction, and (4) persistent lesion identity across repeated observations --- and ran end to end on four public videos (two C3VDv2 sequences with ground-truth depth and two full REAL-Colon procedures; 40,245 frames). All components are published, individually validated methods; the contribution is their lesion-centered assembly, linking rules, and evaluation. \textbf{Results}: Revisits, impossible under forward-only mapping by construction, were detected by entry-map Bayesian localization: 5,614 and 4,043 revisit events (56 and 68 distinct nodes) in the two full procedures. Lesion-identity merging at the adopted threshold 0.5 maintained ground-truth purity 1.0 while auto-merging 20 of 231 candidate pairs. The endoscopy-specific geometry engine outperformed a general-purpose foundation model on all metrics (overall absolute relative error (AbsRel) 0.2276 vs. 0.3523). \textbf{Conclusions}: The results are partial but establish a concrete near-term path: revisit detection, lesion identity, and local 3D each returned quantitative, reproducible output without waiting for complete geometric reconstruction; validating the record on clinical data is the next step.
\end{abstract}


\end{frontmatter}

\section{Introduction}\label{sec:intro}

Colorectal cancer remains among the most incident cancers worldwide~\cite{ref:bray}, and colonoscopy is the central examination for screening, surveillance, and endoscopic resection~\cite{ref:rex}. During a procedure, the endoscopist traverses the colonic segments during insertion and withdrawal, detects lesions, and observes, resects, and records them. What persists after the procedure is a narrative findings description and a small number of still photographs --- the documentation form codified in colonoscopy quality-indicator guidance~\cite{ref:rex}. The recorded video itself carries where the scope passed, where and from which viewing angle a lesion was observed, and how many times the same lesion reappeared --- but this spatial information is discarded at the end of the examination.

The gap is a recording-culture problem before it is a technology problem. Pathology results persist as standardized reports, whereas lesion location depends on narrative descriptions such as ``distal transverse colon,'' and inaccurate preoperative localization of colonic neoplasia is a documented source of surgical management error~\cite{ref:fernandez}. Locating a previously treated lesion at surveillance colonoscopy relies on the earlier video and the physician's memory, which makes between-examination comparison and longitudinal lesion tracking difficult~\cite{ref:rex}. A fraction of lesions is additionally missed even within a single examination~\cite{ref:zhao}, so the evidentiary record that video could provide is lost precisely where completeness matters.

The technical landscape is nonetheless ready to answer this question: public collections now provide complete calibrated procedures~\cite{ref:endomapper}, real multi-center recordings~\cite{ref:realcolon}, and phantom sequences with dense ground truth~\cite{ref:c3vd}. Recent work in endoscopic 3D reconstruction and place recognition has advanced rapidly, but each line of research pursued a different goal, and no study has asked what these components deliver when assembled around lesions. This study addresses that question: under realistic conditions where tracking breaks frequently~\cite{ref:cudaSIFT} and tissue deforms~\cite{ref:defslam}, can a 3D map that preserves (1) which segments were traversed, (2) where lesions were observed, and (3) whether the same lesion is being seen again be produced without reconstructing the whole colon as a single precise 3D model?

Several research threads lead to this position. Rigid-motion visual simultaneous localization and mapping (SLAM) fails in real procedures, where peristalsis, haustral-fold occlusion, washing fluid, specular reflections, and fast camera motion break feature tracking~\cite{ref:cudaSIFT}. Benchmark studies on ex-vivo and synthetic data report the same failure modes~\cite{ref:endoslam}, and deformable tissue motivated dedicated non-rigid SLAM formulations~\cite{ref:defslam,ref:nrslam}. CudaSIFT-SLAM~\cite{ref:cudaSIFT} countered this with GPU SIFT features and a multi-map backend that restarts a submap at every tracking loss and merges submaps over common views; it is one of the few systems to process a full real procedure in real time, yet its authors report that close observation covered only about 38\% of the procedure --- evidence both of how difficult complete millimeter-scale mapping is and of the need for a higher structure that manages submap fragmentation. ColonSLAM~\cite{ref:colonslam} changed the objective from complete geometry to a graph of places, combining global descriptors from a place-recognition network with order priors (the endoscope does not travel backward in the colon) to organize an entire procedure as one topological graph; its verification stage uses LightGlue matching~\cite{ref:lightglue}. ColonMapper~\cite{ref:colonmapper} extended place recognition across examinations with deeply learned global descriptors and a Bayesian filter, and its two-phase protocol --- build the map during insertion, localize withdrawal frames against that map --- provides the conceptual bridge between within-procedure revisits and across-examination alignment.

On geometry, monocular depth estimation has progressed from single-frame networks to temporally consistent streaming models~\cite{ref:endostreamdepth} and to densification approaches that scale-align dense depth to sparse SLAM submaps~\cite{ref:densification}; integrated online foundation models that output pointmaps, depth, and camera parameters~\cite{ref:endo3r}, together with general-purpose geometry transformers such as VGGT~\cite{ref:vggt}, now offer alternatives to hand-assembled modules. Related pipelines such as PERSEUS~\cite{ref:perseus} and Semantic-SuPer~\cite{ref:semanticsuper} demonstrate semantic 3D reconstruction in endoscopic scenarios. On lesion identity, SALI established the view that lesions are spatio-temporal objects rather than per-frame detections, validated on the large-scale SUN-SEG video polyp segmentation benchmark~\cite{ref:sali,ref:vps}, and promptable segmentation models such as SAM~2 propagate an operator click or box through the entire video~\cite{ref:sam2}. What remains beyond tracking is identity: when propagation loses the target the masklet ends, so a re-observed lesion starts a new track~\cite{ref:sam2}, and deciding whether two tracks show the same lesion becomes a separate problem~\cite{ref:sali}.

This study contributes the following. First, it organizes these separately developed components --- endoscopic SLAM, topological mapping, depth estimation, video lesion tracking --- from a lesion-centered mapping perspective, and states explicitly what each solved and what each left open. Second, it proposes a four-layer hierarchical ensemble (global topological map, lesion tracks, selective local 3D, persistent lesion ID) with defined inter-layer linking rules, and runs the entire pipeline end to end on four public videos, reporting what holds and what does not. Third, it identifies the conditions for clinical adoption and the remaining problems, framing the algorithmic contributions planned as follow-up work. The paper reports the implementation and validation of this ensemble on public benchmarks; it does not claim new base algorithms.

\section{Materials and Methods}\label{sec:methods}

\subsection{Lesion-centered hierarchical ensemble pipeline}\label{sec:pipeline}

\begin{figure}[t]
\centering
\includegraphics[width=\linewidth]{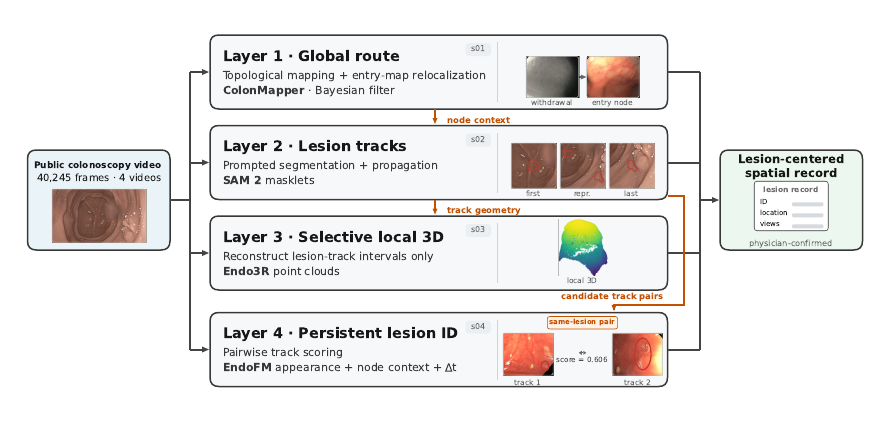}
\caption{Lesion-centered hierarchical ensemble pipeline. Four layers --- global topological map, lesion tracks, selective local 3D, and persistent lesion identity --- with the inter-layer linking rules (node context, track geometry, candidate track pairs). Each layer carries one representative output: a re-localization pair (withdrawal frame and matched entry-map node, REAL-Colon), the first/representative/last frames of one lesion track with segmentation contours, a local 3D point cloud, and a ground-truth-verified pair of fragmented tracks of the same lesion, compared by pairwise scoring with the lesion contoured in both views (REAL-Colon); the output card sketches the record fields with placeholder bars.}\label{fig:pipeline}
\end{figure}

The proposed framework takes a recorded procedure video as input and processes it in four layers (Fig.~\ref{fig:pipeline}). The first layer is the \emph{global topological map}: topological mapping in the ColonMapper family~\cite{ref:colonmapper} organizes the procedure route into a node graph, and a Bayesian localization phase assigns withdrawal-phase frames to the insertion-phase map to detect revisits. The second layer is \emph{lesion tracking}: SAM~2~\cite{ref:sam2} is seeded from an operator box (or dataset frame boxes) and propagates masklets, producing lesion-level spatio-temporal tracks with representative frames. The third layer is \emph{selective local 3D}: point clouds are computed with Endo3R~\cite{ref:endo3r} only over intervals where tracks exist, not over the whole procedure. The fourth layer is \emph{persistent lesion identity}: track pairs are scored by combining appearance embeddings of representative frames (EndoFM~\cite{ref:endofm}) with node co-assignment and temporal separation, and each pair is classified into auto-merge, review queue, or reject.

The boundary of the contribution is stated explicitly. Each layer's components are public implementations of existing research; the contribution is the layered structure with its registry contracts and the inter-layer linking rules --- node context of tracks, track geometry as the trigger for selective reconstruction, and the pairwise identity score.

\subsection{Data}\label{sec:data}

Four public videos with complementary characteristics were selected (Table~\ref{tab:data}). The two C3VDv2 sequences~\cite{ref:c3vd,ref:c3vdv2} are realistic phantom recordings accompanied by ground-truth depth and serve as the quantitative reference for local 3D (tier t1); the cecum sequence provides the depth reference point and the transverse sequence adds a deforming, non-rigid condition. The two REAL-Colon videos~\cite{ref:realcolon} are full real procedures and serve as integrated cases for topology, revisit detection, and identity (tier t2); one of them contains a long re-observation gap. The total expansion is 40,245 frames.

\begin{table}[t]
\caption{Experiment data --- four public videos}\label{tab:data}
\footnotesize\setlength{\tabcolsep}{4pt}
\begin{tabular}{@{}llll@{}}
\toprule
Slot & Dataset, video & Frames (expanded) & Role \\
\midrule
c3vdv2-cecum & C3VDv2 c1\_cecum\_t1\_v2 & 423 (stride 1) & t1: GT-depth reference \\
c3vdv2-transverse & C3VDv2 c2\_transverse1\_t4\_v3 & 382 (stride 1) & t1: non-rigid condition \\
realcolon-1 & REAL-Colon 001-004 & 45,663 $\to$ 22,832 (stride 2) & t2: full real procedure \\
realcolon-2 & REAL-Colon 002-008 & 33,216 $\to$ 16,608 (stride 2) & t2: re-observation gap \\
\bottomrule
\end{tabular}
\end{table}

Lesion prompts were operator boxes for t1 and REAL-Colon frame boxes in PASCAL VOC XML format for t2. Ground-truth (GT) lesion identifiers and ground-truth depth were used for evaluation only and never entered any pipeline decision --- the leakage prohibition is maintained from experiment design through reporting. The pipeline ran in seven stages (frame expansion, topology, lesion tracking, geometry, identity, reporting, quantitative evaluation) and completed in about 4.3~hours on a single GPU.

\subsection{Evaluation protocols}\label{sec:protocols}

\emph{Revisit detection.} Following the two-phase protocol of ColonMapper~\cite{ref:colonmapper}, the map is cut at the end of insertion (entry map; cut at the median registry index) and deployed frames are localized against it. The final configuration uses the authors' default gate of 0.5~\cite{ref:colonmapper} (entry\_tp050); a laxer gate of 0.33 and a full-map variant are reported alongside for comparison. A revisit event is a frame-to-node assignment whose temporal gap exceeds 150 frames.

\emph{Lesion identity.} Candidate track pairs from the integrated run are scored, and an auto-merge threshold grid is evaluated against ground-truth lesion identity: the number of auto-merged pairs, review-queue size, rejects, ground-truth purity of merged clusters, and same-lesion recall. The default threshold 0.75 and the adopted threshold 0.5 are reported side by side, with the reject threshold fixed at 0.35.

\emph{Local 3D.} On the two t1 sequences, 32 frames per video (linearly spaced) are evaluated with per-frame median scale alignment against ground-truth depth over valid ground-truth pixels, reporting absolute relative error (AbsRel), root-mean-square error in millimeters (RMSE), and the fraction of pixels with relative error below 0.25 ($\delta<1.25$). The endoscopy-specific online engine Endo3R~\cite{ref:endo3r} is compared with the general-purpose foundation model VGGT~\cite{ref:vggt}.

\emph{Qualitative track quality.} All tracks are classified by a rule-based proxy (blur from a sharpness percentile, washing from a specular-highlight percentile, drift from mask-area discontinuities, and local-3D feasibility from point-cloud scores and extent), with the threshold constants recorded in the released classification artifacts for reproducibility.

\section{Results}\label{sec:results}

\subsection{Revisit detection}\label{sec:revisit}

\begin{figure}[t]
\centering
\includegraphics[width=\linewidth]{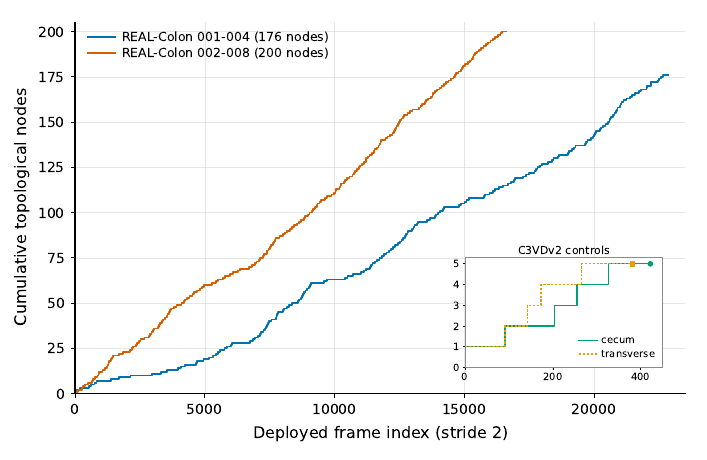}
\caption{Forward-only node growth of the mapping phase. The node count increases monotonically with procedure progress and never returns --- the structural reason why mapping alone yields zero revisits.}\label{fig:nodegrowth}
\end{figure}

A structural fact comes first. Topological mapping grows only forward and compares each incoming frame only with the current proto-node, so revisits cannot arise in the mapping output by construction~\cite{ref:colonmapper,ref:colonslam}. This was confirmed empirically: in both full procedures (176 and 200 nodes), the mapping phase alone produced zero revisit nodes. Revisit detection is the responsibility of the localization phase (Fig.~\ref{fig:nodegrowth}).

\begin{table}[t]
\caption{Revisit detection --- mapping alone vs. entry-map localization (entry\_tp050, final run)}\label{tab:revisit}
\begin{tabular}{@{}llllll@{}}
\toprule
Video & Method & Localized & Events ($>$150) & Nodes & Max gap \\
\midrule
realcolon-1 & mapping only & -- & 0 & 0 & -- \\
realcolon-1 & entry\_tp050 & 0.8011 & 5,614 & 56 & 22,203 \\
realcolon-2 & mapping only & -- & 0 & 0 & -- \\
realcolon-2 & entry\_tp050 & 0.7794 & 4,043 & 68 & 15,630 \\
c3vdv2-cecum (control) & entry\_tp050 & 1.0 & 56 & 1 & 261 \\
c3vdv2-transverse (control) & entry\_tp050 & 1.0 & 34 & 1 & 218 \\
\bottomrule
\end{tabular}

\raggedright\small Gaps are in registry-frame units; a revisit requires a gap $>$ 150. The t1 controls have small graphs (3--5 nodes) and only a single revisit node, separating clearly from the t2 procedures.
\end{table}

\begin{figure}[t]
\centering
\includegraphics[width=\linewidth]{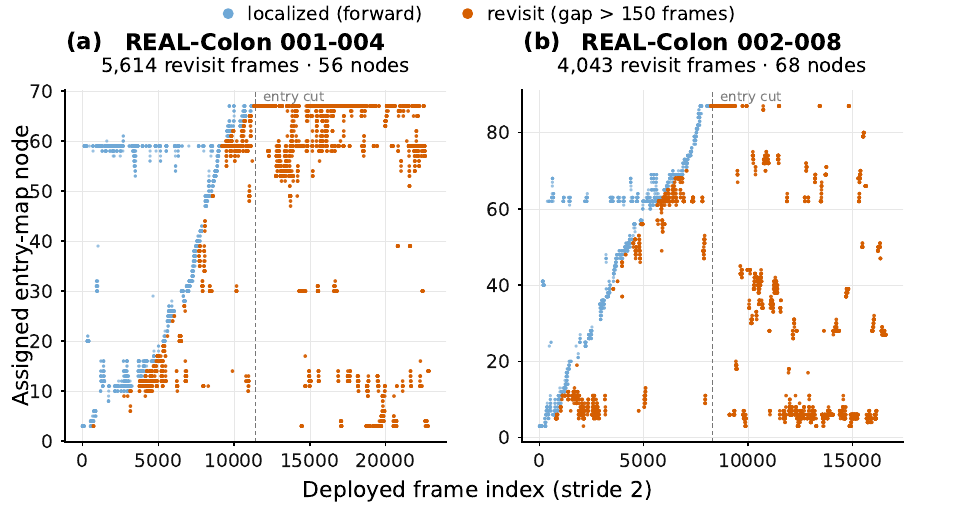}
\caption{Frame-to-entry-map node assignment over the full procedure ((a) REAL-Colon 001-004, (b) REAL-Colon 002-008); the dashed line marks the entry cut at the end of insertion. Vermillion points are revisit-flagged assignments whose temporal gap exceeds 150 frames --- insertion-phase re-localization jumps as well as withdrawal re-passages. The assignment of the late-withdrawal tail to early insertion nodes is physically consistent with re-passage through the rectum.}\label{fig:assignment}
\end{figure}

\begin{figure}[t]
\centering
\includegraphics[width=\linewidth]{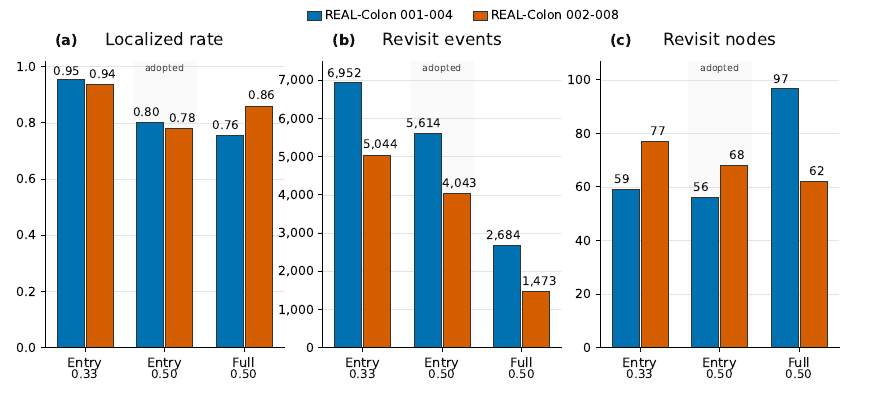}
\caption{Revisit-detection variant comparison: entry-map vs.\ full-map localization at gate thresholds 0.33 and 0.50 (a revisit event has a temporal gap $>$ 150 frames). The grey band marks the adopted configuration (entry map, gate 0.50). Cutting the map at the end of insertion yields substantially more revisit events than the full map (5,614 vs.\ 2,684 and 4,043 vs.\ 1,473), supporting the two-phase protocol of ColonMapper~\cite{ref:colonmapper}.}\label{fig:sweep}
\end{figure}

Table~\ref{tab:revisit} and Figs.~\ref{fig:assignment}--\ref{fig:sweep} summarize the outcome. The largest gaps (15,630 and 22,203 frames) correspond to end-of-withdrawal frames assigned to the earliest insertion nodes, which agrees with the endoscope re-passing the rectum (Fig.~\ref{fig:assignment}). The t1 controls were insensitive to the gate (identical results at gates 0.33 and 0.5), whereas on t2 the laxer gate 0.33 raised the localized fraction to 0.94--0.95 at the cost of possible over-assignment; the reported configuration therefore adopts the authors' default gate 0.5~\cite{ref:colonmapper}, with all variants disclosed. Entry-map cutting produced 2.1--2.7$\times$ more events than full-map localization, confirming the advantage of the entry-map protocol (Fig.~\ref{fig:sweep}).

\subsection{Lesion identity}\label{sec:identity}

\begin{figure}[t]
\centering
\includegraphics[width=\linewidth]{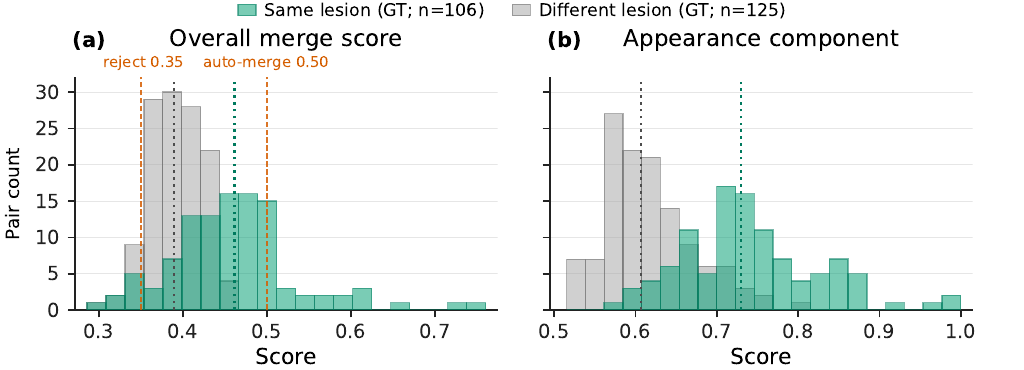}
\caption{Raw merge-score distributions over 231 candidate pairs --- final score (a) and appearance component (b). Green: same ground-truth lesion (106 pairs); gray: different lesions (125). Dotted lines are medians; vermillion dashed lines are the reject 0.35 and auto-merge 0.50 thresholds. The distributions overlap broadly, and only the appearance component contributes to separation.}\label{fig:components}
\end{figure}

The identity target is 231 candidate pairs (106 same-lesion, 125 different-lesion by ground truth). Score medians are 0.4612 for same-lesion and 0.3895 for different-lesion pairs --- separation exists but is narrow (Fig.~\ref{fig:components}).

\begin{table}[t]
\caption{Lesion identity --- default threshold 0.75 vs. adopted 0.5 (231 candidate pairs, reject fixed at 0.35)}\label{tab:identity}
\begin{tabular}{@{}lllllll@{}}
\toprule
Auto-merge & Merged & Queue & Rejected & GT-mixed & Purity & Same-GT recall \\
\midrule
0.75 (default) & 1 & 214 & 16 & 0 & 1.0 & 0.0094 \\
0.5 (adopted) & 20 & 195 & 16 & 0 & 1.0 & 0.1887 \\
0.45 & 61 & 154 & 16 & 1 & 0.9836 & 0.566 \\
0.40 & 141 & 74 & 16 & 2 & 0.617 & 0.8208 \\
\bottomrule
\end{tabular}

\raggedright\small The adopted threshold 0.5 is the largest value that keeps purity 1.0 (first mixing at 0.45, collapse at 0.40). Under the default threshold, cluster counts per ground-truth lesion are 6/7/10/1/6 --- a fragmented, conservative behavior.
\end{table}

\begin{figure}[t]
\centering
\includegraphics[width=\linewidth]{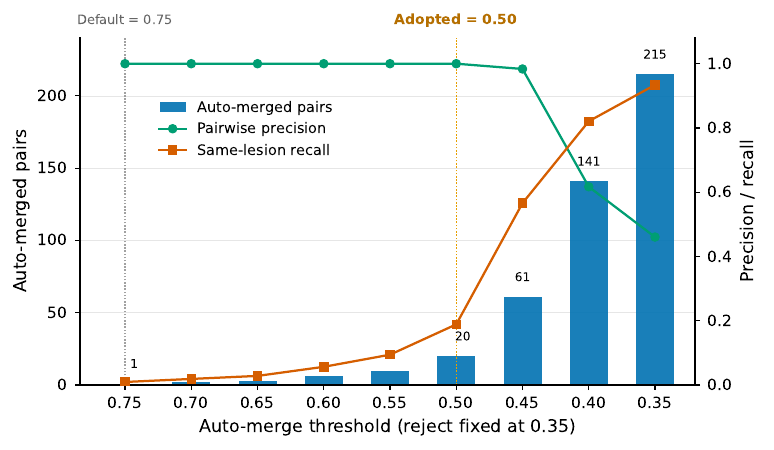}
\caption{Auto-merge threshold grid --- number of auto-merged pairs (bars) and ground-truth precision and recall (lines). The adopted threshold 0.5 is the largest value that maintains precision 1.0.}\label{fig:threshold}
\end{figure}

The threshold analysis (Table~\ref{tab:identity}, Fig.~\ref{fig:threshold}) shows that 0.5 is the largest threshold that keeps purity at 1.0; below it, mixing appears at 0.45 and collapses at 0.40. The 20 auto-merged pairs at the adopted threshold contain no ground-truth mixing, so proposals contributed only to shrinking the review queue. The node component of the score is zero throughout --- under forward-only mapping, node co-assignment between tracks from different times cannot occur --- and this absence is the structural origin of the recall ceiling of 0.19. The distributions (Fig.~\ref{fig:components}) show broad overlap with only the appearance component contributing to separation. Under the default threshold, ground-truth lesions fragment into 6/7/10/1/6 clusters: repeated observations of the same lesion remain separate, a conservative behavior documented as a limitation.

\subsection{Local 3D}\label{sec:local3d}

\begin{figure}[t]
\centering
\includegraphics[width=\linewidth]{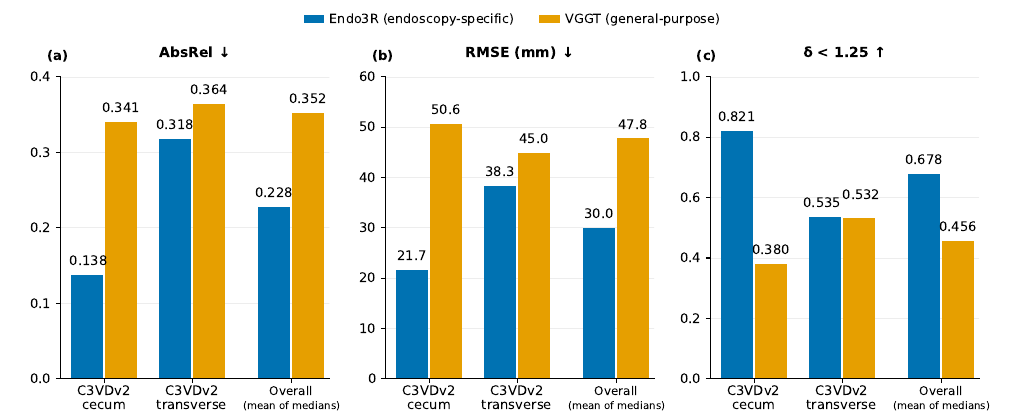}
\caption{Local 3D quantitative comparison on the t1 videos (Endo3R vs. VGGT): (a) AbsRel, (b) RMSE (mm), and (c) $\delta<1.25$ over 32 frames per video with per-frame median scale alignment. Overall denotes the mean of the per-video medians.}\label{fig:depth}
\end{figure}

On the t1 quantitative comparison (Table~\ref{tab:depth}, Fig.~\ref{fig:depth}), the endoscopy-specific online engine (Endo3R~\cite{ref:endo3r}) outperformed the general-purpose foundation model (VGGT~\cite{ref:vggt}) on both videos and on all summary metrics; on the transverse sequence the $\delta<1.25$ values are effectively tied (0.535 vs. 0.5324).

\begin{table}[t]
\caption{Local 3D quantitative results --- Endo3R vs. VGGT (32 frames per video, per-frame median scale alignment)}\label{tab:depth}
\begin{tabular}{@{}lllll@{}}
\toprule
Engine & Video & AbsRel & RMSE (mm) & $\delta<1.25$ \\
\midrule
Endo3R & c3vdv2-cecum & 0.1376 & 21.66 & 0.8208 \\
Endo3R & c3vdv2-transverse & 0.3175 & 38.33 & 0.535 \\
Endo3R & overall (mean of medians) & 0.2276 & 29.99 & 0.6779 \\
VGGT & c3vdv2-cecum & 0.3408 & 50.58 & 0.3804 \\
VGGT & c3vdv2-transverse & 0.3638 & 44.96 & 0.5324 \\
VGGT & overall (mean of medians) & 0.3523 & 47.77 & 0.4564 \\
\bottomrule
\end{tabular}

\raggedright\small The transverse $\delta<1.25$ values are effectively tied; Endo3R leads on all remaining metrics.
\end{table}

The weakness on the transverse sequence is structural: a mid-sequence abrupt scale drop and a sharp AbsRel increase over the last five frames were observed, indicating the limit of online monocular depth under strong non-rigid deformation, the difficulty that motivates non-rigid formulations~\cite{ref:defslam,ref:nrslam}.

\subsection{Qualitative behavior}\label{sec:qualitative}

\begin{figure}[t]
\centering
\includegraphics[width=\linewidth]{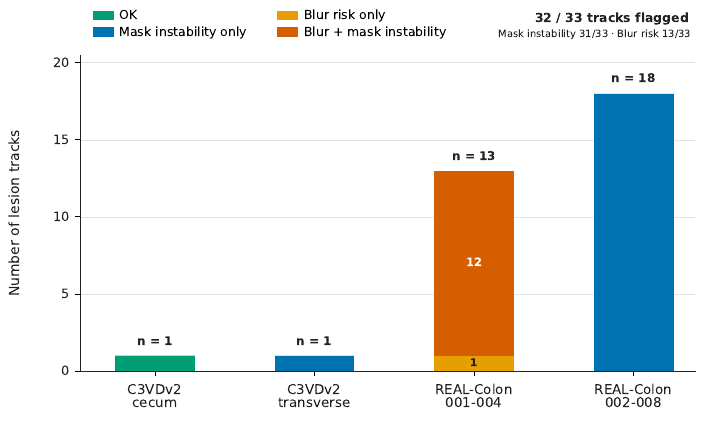}
\caption{Track quality flag distribution (rule-based proxy classification).}\label{fig:quality}
\end{figure}

\begin{figure}[t]
\centering
\includegraphics[width=\linewidth]{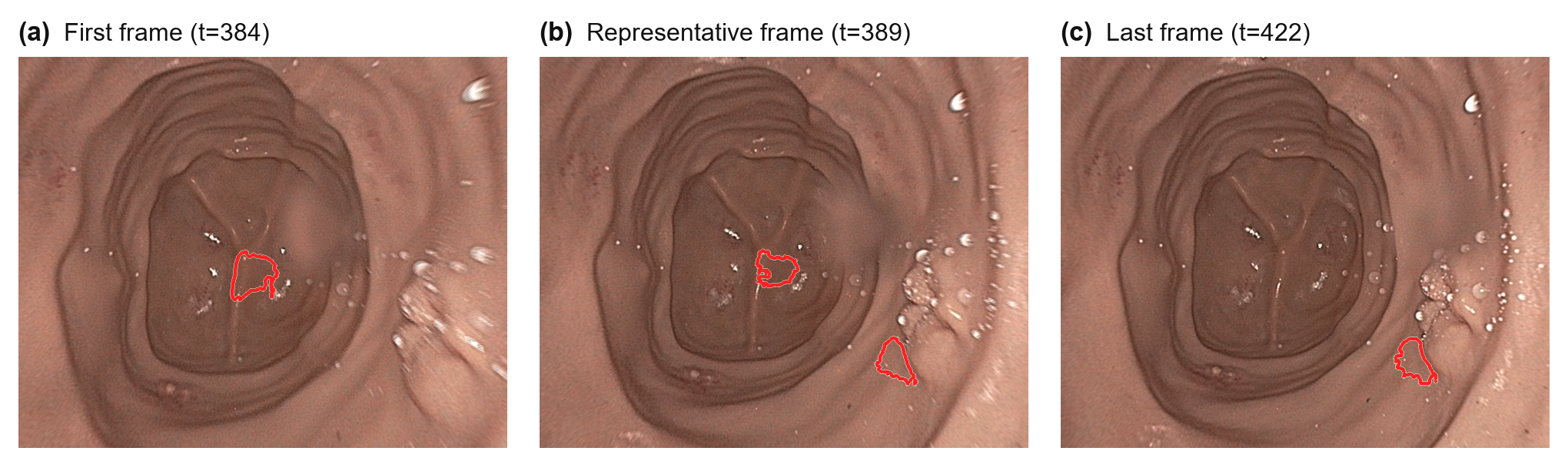}
\caption{Lesion track examples on C3VDv2 (t1) --- SAM~2~\cite{ref:sam2} mask outlines in red (two disconnected regions on the representative frame).}\label{fig:trackst1}
\end{figure}

\begin{figure}[t]
\centering
\includegraphics[width=0.92\linewidth]{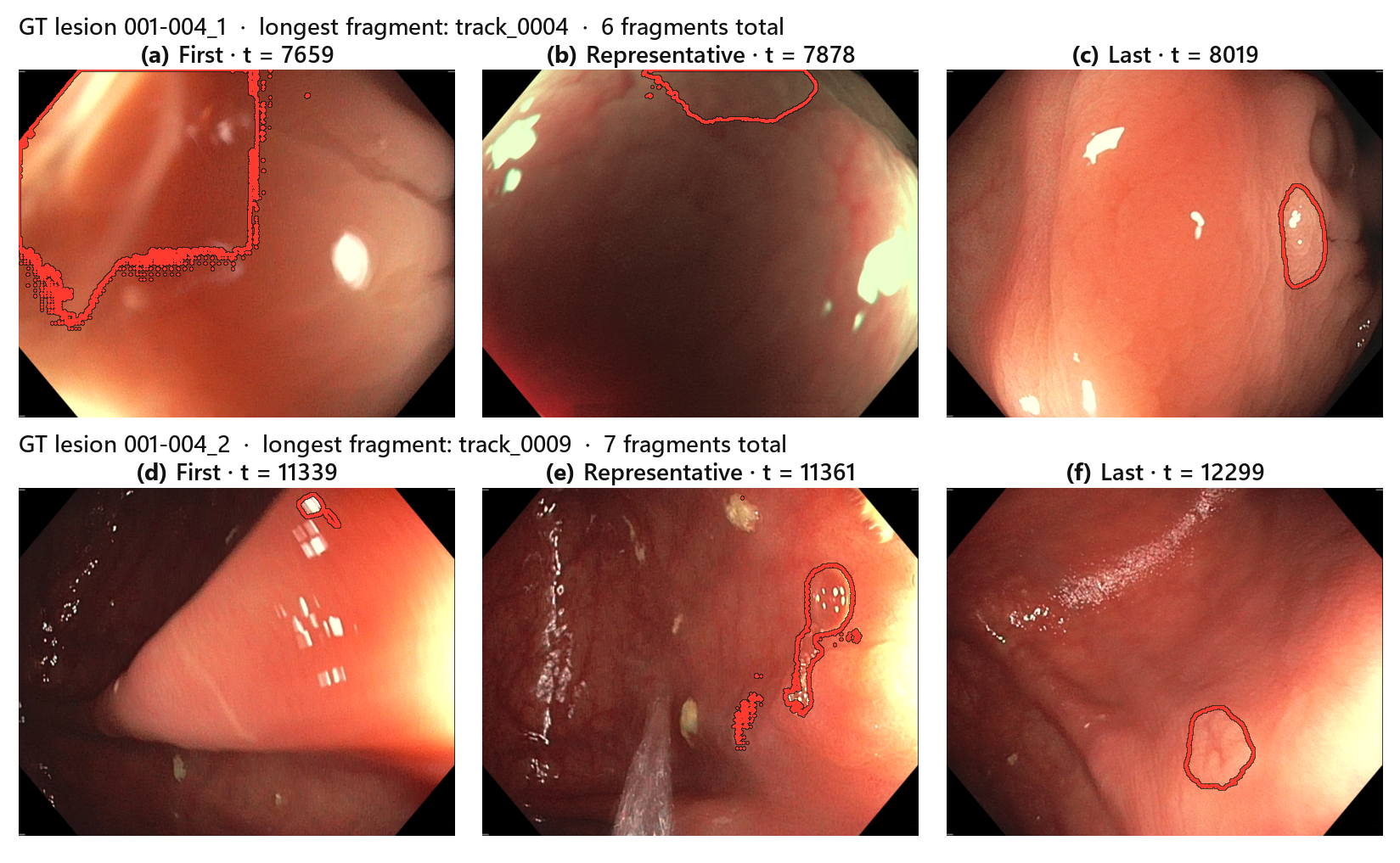}
\caption{Lesion track examples on REAL-Colon (t2) --- the longest track per ground-truth lesion, shown together with fragmentation.}\label{fig:trackst2}
\end{figure}

\begin{figure}[t]
\centering
\includegraphics[width=0.88\linewidth]{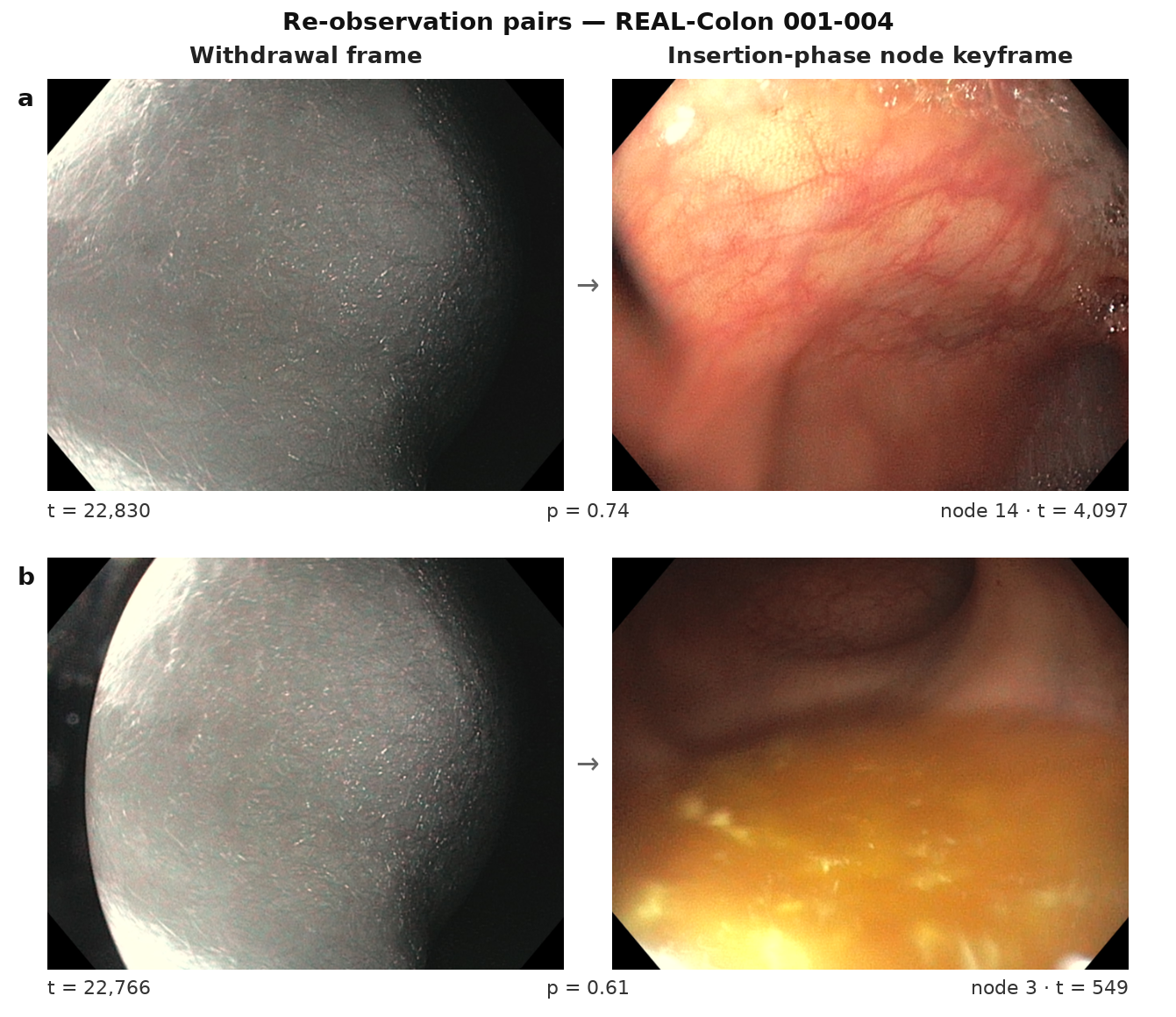}
\caption{Re-observation frame pairs --- (a) and (b) are algorithmically re-localized matched pairs, each pairing a withdrawal-phase query frame (left) with the insertion-phase node keyframe to which it was re-localized (right). $p$ is the Bayesian node-assignment posterior probability under the adopted entry-map gate of 0.5. No photometric modification was applied to the frames. The figure is a qualitative example of localization output, not a claim of anatomical same-place ground truth.}\label{fig:pairs}
\end{figure}

\begin{figure}[t]
\centering
\includegraphics[width=0.72\linewidth]{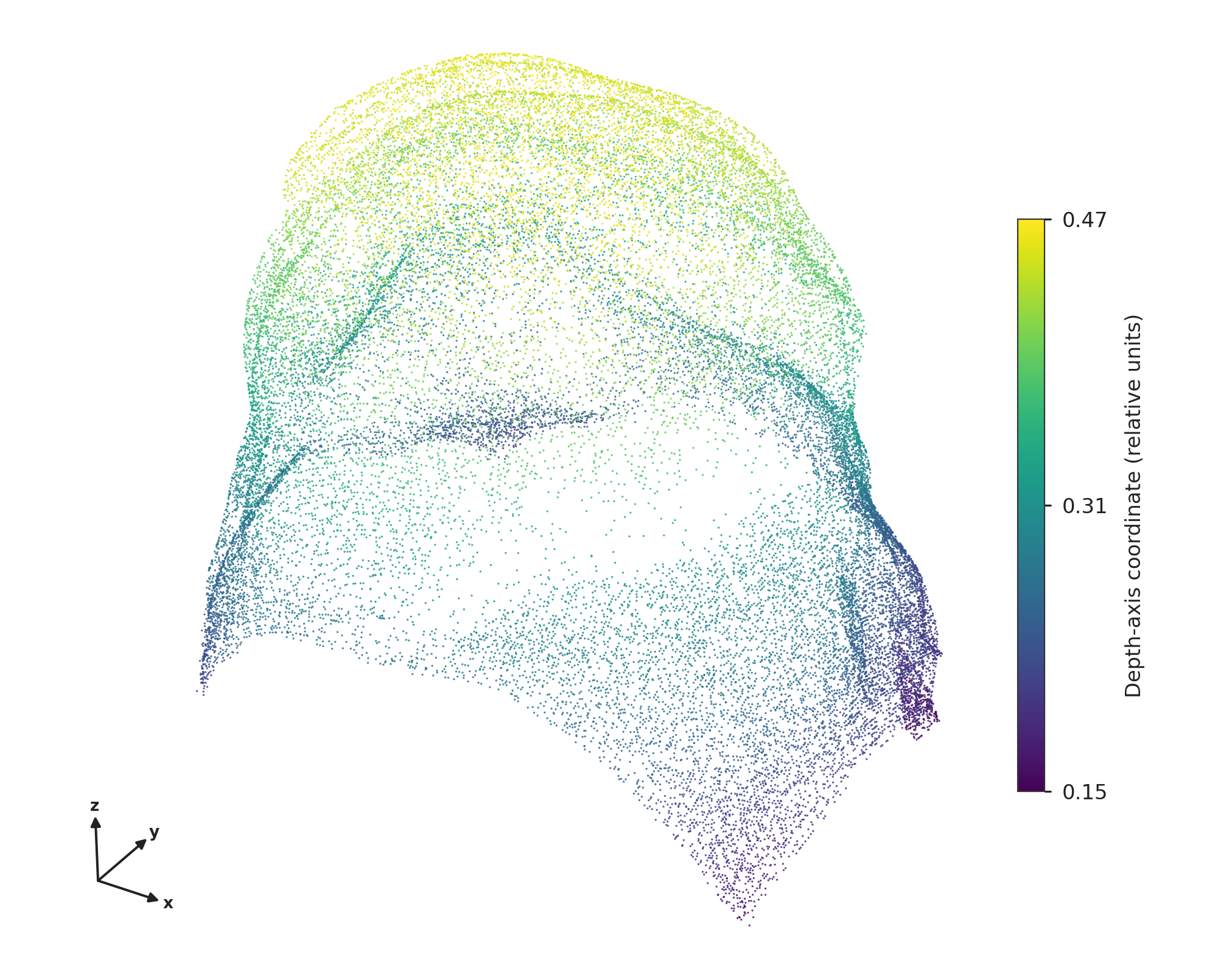}
\caption{Selective local 3D reconstruction over a lesion-track interval in the C3VDv2 cecum sequence using Endo3R~\cite{ref:endo3r}. Points are colored by the depth-axis coordinate in relative units. The reconstruction is shown as relative geometry rather than metric scale.}\label{fig:ply}
\end{figure}

Rule-based quality classification of the 33 tracks yielded one clean track and 32 flagged tracks (mask instability 31, blur risk 13) (Fig.~\ref{fig:quality}). Mask instability dominates on REAL-Colon while the C3VDv2 cecum track is clean --- the difficulty of real procedures is reflected directly. Figures~\ref{fig:trackst1}--\ref{fig:ply} show lesion-track examples on both tiers, re-observation frame pairs that ground the revisit detections of Section~\ref{sec:revisit}, and a local 3D point cloud produced by the selective reconstruction layer.

\subsection{What the validation establishes}\label{sec:lessons}

First, layer separation creates a structure in which each question becomes answerable: zero revisits under mapping alone is not a failure but a structural fact, and attaching the localization phase surfaced thousands of revisit events in the same videos. Second, weak inter-layer linking limits identity: the zero node component and the resulting recall ceiling expose the missing connection between forward-only mapping and the identity score. Third, the difficulty is uneven across layers --- local 3D is already usable on phantom data, whereas mask instability and non-rigid deformation in real procedures remain open challenges for both the track and geometry layers~\cite{ref:sam2,ref:nrslam}.

\section{Discussion}\label{sec:discussion}

\subsection{Clinical relevance of lesion-centered maps}

A lesion-centered map offers three values: during look-back exploration it provides the observation positions and viewpoints of the previous examination as spatial context; during surveillance it creates continuity of lesion-level records~\cite{ref:fernandez}; and for documentation standardization it adds a spatial axis to procedure records~\cite{ref:rex}. Missed lesions remain a measured concern in practice~\cite{ref:zhao}, and a record that preserves which segments were traversed and in what order is a prerequisite for auditing completeness. In this validation, both revisit detection and auto-merging operated at the level of candidate proposal, which matches the confirmation-loop structure in which the final decision rests with the physician.

\subsection{The gap between public data and clinical data}

The quality difference between the phantom tier and the real-procedure tier reflects that public benchmarks do not fully represent clinical imaging difficulty~\cite{ref:c3vd,ref:realcolon,ref:endomapper}: mask instability in 31 of 33 tracks means SAM~2~\cite{ref:sam2} propagation breaks far more often in real procedures, which directly affects the volume and fragmentation of identity candidates. Clinical adoption assessment requires validation on real procedure data, together with data acquisition and de-identification procedures. This gap also has an evaluation-methodology dimension: auditing work on AI benchmarks shows that reliable assessment requires checking the task, environment, ground truth, and evaluation components themselves, not only model outputs~\cite{ref:agentsuite}; the leakage prohibition, threshold-variant disclosure, and run-level reproducibility of this study follow the same principle.

\subsection{Limitations and ethical position}

The scale of monocular depth is intrinsically uncertain --- the motivation for scale-consistent endoscopic reconstruction~\cite{ref:endo3r} --- and the local 3D of this framework must be read as relative structure. Non-rigid deformation, a standing difficulty for monocular endoscopy~\cite{ref:nrslam}, degrades end-of-sequence accuracy, as the transverse case showed. Above all, every automated decision is a proposal, and the physician remains the final authority for confirming lesion records. The scale of this validation (four videos) is directional; no generalization claim is made.

\subsection{Future work}

Four directions follow directly. (1) Feed localization-phase re-assignments back into the node component of the identity score to relax the recall ceiling. (2) Compare non-rigid formulations and dynamic scene representations (e.g., NR-SLAM~\cite{ref:nrslam}, ColonSplat~\cite{ref:colonsplat}) as alternatives for the local 3D layer. (3) Extend persistent lesion identity across examinations via cross-procedure alignment. (4) Replace operator prompts with semi-automatic detection to reduce operator burden. These directions are the algorithmic contributions planned as follow-up work.

\section{Conclusion}\label{sec:conclusion}

To address the loss of spatial information in colonoscopy recording practice, this study proposed a lesion-centered hierarchical ensemble as an alternative to complete 3D reconstruction and ran it end to end on four public videos. Revisit detection answered with 5,614 and 4,043 events where mapping alone yields zero; lesion identity auto-merged 20 pairs at a threshold that keeps purity 1.0; and local 3D favored the endoscopy-specific engine on every metric. The results are partial but show that preserving the spatial context already present in procedure video, centered on lesions, is a genuinely open path. Strengthening the inter-layer links and validating on clinical data remain the tasks ahead.

\section*{CRediT authorship contribution statement}
\textbf{Hyunjun Kim:} Conceptualization, Methodology, Software, Investigation, Data curation, Writing -- original draft.
\textbf{Hyeonwoo Na:} Conceptualization, Investigation, Data curation, Writing -- review \& editing.
\textbf{Jaewoo Lee:} Conceptualization, Writing -- review \& editing.

\section*{Declaration of competing interest}
The authors declare that they have no known competing financial interests or personal relationships that could have appeared to influence the work reported in this paper.

\section*{Funding}
This research did not receive any specific grant from funding agencies in the public, commercial, or not-for-profit sectors.

\section*{Ethics statement}
This study used only publicly available, de-identified benchmark datasets (C3VDv2 and REAL-Colon). No participants were recruited and no new human data were collected by the authors; no ethics committee approval was therefore sought because no human participants or personal data were involved.

\section*{Data availability}
The datasets analyzed are publicly available from their original publishers (C3VDv2~\cite{ref:c3vdv2}; REAL-Colon~\cite{ref:realcolon}). Other research artifacts are available from the corresponding author on reasonable request.

\section*{Code availability}
The complete pipeline code, adapters, experiment configurations, and test suite are openly available at \url{https://github.com/hyunjun1121/endovision-pipeline} (Zenodo DOI \url{https://doi.org/10.5281/zenodo.22136766}).

\section*{Declaration of generative AI and AI-assisted technologies in the writing process}
During the preparation of this work the author(s) used ZCode, an AI coding assistant powered by the GLM large language model (Z.ai), in order to assist with drafting and language editing of the manuscript text. After using this tool/service, the author(s) reviewed and edited the content as needed and take(s) full responsibility for the content of the published article.

\appendix

\section{Reproducibility note}\label{sec:appendix}

The complete run is reproducible from a single configuration file, with the run snapshot and validation report stored alongside the artifacts (base-layer comparison run kci-4v-s03-001; final run kci-4v-s04-001). All quantitative figures (Figs.~\ref{fig:assignment}--\ref{fig:quality}) are generated automatically from repository result JSON files, and the qualitative assets are extracted from the same run artifacts. Dataset licenses: C3VDv2, CC BY 4.0~\cite{ref:c3vdv2}; REAL-Colon, CC BY-NC-SA 4.0~\cite{ref:realcolon}. The complete pipeline code, adapters, configurations, and test suite are openly available (see the Code availability statement).



\begin{thebibliography}{27}

\bibitem{ref:bray} Bray F, Laversanne M, Sung H, Ferlay J, Siegel RL, et al: Global cancer statistics 2022: GLOBOCAN estimates of incidence and mortality worldwide for 36 cancers in 185 countries. CA Cancer J Clin 74:229-263, 2024

\bibitem{ref:rex} Rex DK, Anderson JC, Butterly LF, Day LW, Dominitz JA, et al: Quality indicators for colonoscopy. Gastrointest Endosc 100:352-381, 2024

\bibitem{ref:fernandez} Fernandez LM, Ibrahim RNM, Mizrahi I, DaSilva G, Wexner SD: How accurate is preoperative colonoscopic localization of colonic neoplasia? Surg Endosc 33:1174-1179, 2019

\bibitem{ref:zhao} Zhao S, Wang S, Pan P, Xia T, Chang X, et al: Magnitude, risk factors, and factors associated with adenoma miss rate of tandem colonoscopy: a systematic review and meta-analysis. Gastroenterology 156:1661-1674, 2019

\bibitem{ref:endomapper} Azagra P, Sostres C, Ferr\'andez \'A, Riazuelo L, Tomasini C, et al: EndoMapper dataset of complete calibrated endoscopy procedures. Sci Data 10:671, 2023

\bibitem{ref:realcolon} Biffi C, Antonelli G, Bernhofer S, Hassan C, Hirata D, et al: REAL-Colon: a dataset for developing real-world AI applications in colonoscopy. Sci Data 11:539, 2024

\bibitem{ref:c3vd} Bobrow TL, Golhar M, Vijayan R, Akshintala VS, Garcia JR, Durr NJ: Colonoscopy 3D video dataset with paired depth from 2D-3D registration. Med Image Anal 90:102956, 2023

\bibitem{ref:cudaSIFT} Elvira R, Tard\'os JD, Montiel JMM: CudaSIFT-SLAM: multiple-map visual SLAM for full procedure mapping in real human endoscopy. arXiv preprint arXiv:2405.16932, 2024

\bibitem{ref:defslam} Lamarca J, Parashar S, Bartoli A, Montiel JMM: DefSLAM: tracking and mapping of deforming scenes from monocular sequences. IEEE Trans Robot 37:291-303, 2021

\bibitem{ref:endoslam} Ozyoruk KB, Gokceler GI, Bobrow TL, Coskun G, Incetan K, et al: EndoSLAM dataset and an unsupervised monocular visual odometry and depth estimation approach for endoscopic videos. Med Image Anal 71:102058, 2021

\bibitem{ref:nrslam} G\'omez Rodr\'iguez JJ, Montiel JMM, Tard\'os JD: NR-SLAM: nonrigid monocular SLAM. IEEE Trans Robot 40:4252-4264, 2024

\bibitem{ref:colonslam} Morlana J, Tard\'os JD, Montiel JMM: Topological SLAM in colonoscopies leveraging deep features and topological priors. In: Medical Image Computing and Computer-Assisted Intervention -- MICCAI 2024, Lecture Notes in Computer Science, vol 15011. Cham: Springer; 2024, pp. 733-743

\bibitem{ref:lightglue} Lindenberger P, Sarlin PE, Pollefeys M: LightGlue: local feature matching at light speed. In: IEEE/CVF International Conference on Computer Vision (ICCV); 2023, pp. 17627-17638

\bibitem{ref:colonmapper} Morlana J, Tard\'os JD, Montiel JMM: ColonMapper: topological mapping and localization for colonoscopy. In: IEEE International Conference on Robotics and Automation (ICRA); 2024, pp. 6329-6336

\bibitem{ref:endostreamdepth} Li H, Lu D, Wang J, Webster RJ, Oguz I: EndoStreamDepth: temporally consistent monocular depth estimation for endoscopic video streams. In: Medical Imaging with Deep Learning (MIDL), Proceedings of Machine Learning Research 315:1697-1721, 2026

\bibitem{ref:densification} Anad\'on X, Rodr\'iguez-Puigvert J, Montiel JMM: 3D densification for multi-map monocular VSLAM in endoscopy. arXiv preprint arXiv:2503.14346, 2025

\bibitem{ref:endo3r} Guo J, Dong W, Huang T, Ding H, Wang Z, et al: Endo3R: unified online reconstruction from dynamic monocular endoscopic video. In: Medical Image Computing and Computer-Assisted Intervention -- MICCAI 2025; 2025, pp. 170-180

\bibitem{ref:vggt} Wang J, Chen M, Karaev N, Vedaldi A, Rupprecht C, Novotn\'y D: VGGT: visual geometry grounded transformer. In: IEEE/CVF Conference on Computer Vision and Pattern Recognition (CVPR); 2025, pp. 5294-5306

\bibitem{ref:perseus} Acar A, Li F, Stern SS, Al-Zogbi L, Li H, et al: Perseus: perception with semantic endoscopic understanding and SLAM. Int J Comput Assist Radiol Surg, 2026. doi:10.1007/s11548-026-03717-w

\bibitem{ref:semanticsuper} Lin S, Miao AJ, Lu J, Yu S, Chiu ZY, Richter F, Yip MC: Semantic-SuPer: a semantic-aware surgical perception framework for endoscopic tissue identification, reconstruction, and tracking. In: IEEE International Conference on Robotics and Automation (ICRA); 2023, pp. 4739-4746

\bibitem{ref:sali} Hu Q, Yi Z, Zhou Y, Peng F, Liu M, Li Q, et al: SALI: short-term alignment and long-term interaction network for colonoscopy video polyp segmentation. In: Medical Image Computing and Computer-Assisted Intervention -- MICCAI 2024; 2024, pp. 531-541

\bibitem{ref:vps} Ji GP, Xiao G, Chou YC, Fan DP, Zhao K, et al: Video polyp segmentation: a deep learning perspective. Int J Autom Comput 19:531-549, 2022

\bibitem{ref:sam2} Ravi N, Gabeur V, Hu YT, Hu R, Ryali C, et al: SAM 2: segment anything in images and videos. In: International Conference on Learning Representations (ICLR); 2025

\bibitem{ref:endofm} Wang Z, Liu C, Zhang S, Dou Q: Foundation model for endoscopy video analysis via large-scale self-supervised pre-train. In: Medical Image Computing and Computer-Assisted Intervention -- MICCAI 2023; 2023, pp. 101-111

\bibitem{ref:c3vdv2} Golhar MV, Galeano Fretes LS, Ayers L, Akshintala VS, Bobrow TL, Durr NJ: C3VDv2 --- colonoscopy 3D video dataset with enhanced realism. arXiv preprint arXiv:2506.24074, 2025

\bibitem{ref:agentsuite} Suh H, Ji B, Lee S, Khare R, Khan B, et al: AgentSuite: toward more reliable agent evaluation with a component-based benchmark auditing pipeline. OpenReview preprint 2Exmr1eIKZ (ICML 2026 submission), 2026

\bibitem{ref:colonsplat} Smolak-Dy\.zewska W, Kaleta J, Dall'Alba D, Spurek P: ColonSplat: reconstruction of peristaltic motion in colonoscopy with dynamic Gaussian splatting. arXiv preprint arXiv:2603.06860, 2026

\end{thebibliography}
\end{document}